\documentclass[5p, times, twocolumn]{article}
\title{NLS: an accurate and yet easy-to-interpret regression method} 

\author{%
  Victor Coscrato \\
  University College Cork \\
  \texttt{vcoscrato@gmail.com} \\
  \\
  Marco Henrique de Almeida In\'acio \\
Budapest University of Technology and Economics \\
  \texttt{m@marcoinacio.com} \\
  \\
  Tiago Botari \\
  University of S\~{a}o Paulo \\
  \texttt{tiagobotari@gmail.com} \\  
  \\
  Rafael Izbicki \\
  University of S\~{a}o Carlos \\
  \texttt{rafaelizbicki@gmail.com} \\
}

\usepackage[T1]{fontenc}
\usepackage{nameref}
\usepackage[pdfencoding=unicode, psdextra, colorlinks=true, urlcolor=blue, citecolor=blue, linkcolor=blue]{hyperref}
\usepackage{amsfonts, amsmath, amssymb, amsthm, thmtools}
\usepackage{mathrsfs}
\usepackage{cleveref}
\usepackage{indentfirst}
\usepackage{url}
\usepackage{diagbox}
\usepackage{multirow}
\usepackage{tikz}
\usepackage{adjustbox}

\declaretheorem[name=Theorem, refname={Theorem, Theorems}, Refname={Theorem, Theorems}, parent=section]{theorem}

\declaretheorem[name=Example, refname={Example, Examples}, Refname={Example, Examples}, sibling=theorem, style=definition]{example}

\newtheorem{remark}{Remark}
    
\crefname{section}{section}{sections}
\Crefname{section}{Section}{Sections}
\crefname{table}{table}{tables}
\Crefname{table}{Table}{Tables}

\usepackage{color, enumitem, fancyhdr, float, fourier, layout, multirow, setspace, verbatim}
\setlist[enumerate]{leftmargin=*}
\usepackage[colorinlistoftodos]{todonotes}
\usepackage{listings}
\usepackage{caption, subcaption}

\usepackage{algorithm}
\usepackage{algpseudocode}
\renewcommand{\algorithmicrequire}{\textbf{\small Input:}}
\renewcommand{\algorithmicensure}{\textbf{\small Output:}}

\usepackage{natbib}
\usepackage{enumitem}
\usepackage{graphicx}
\graphicspath{{figures/}{../figures/}}

\DeclareMathOperator*{\argmin}{arg\,min}

\def\x{{\vec{x}}}

\def\X{{\vec{X}}}

\def\y{{\vec{y}}}

\def\E{{\textbf{E}}}

\def\P{{\mathbb P}}

\def\t0{{\theta_0}}

\def\1{{\boldmath{1}}}

\usepackage{MnSymbol}

\renewcommand{\vec}[1]{\mathbf{#1}}

\usepackage{pgfplots}
\pgfplotsset{compat=1.14}

\date{}
\usepackage[bottom=2cm, top=2cm, left=2cm, right=2cm]{geometry}

\begin{document}

\twocolumn[
\begin{@twocolumnfalse}
\maketitle
\begin{abstract}
    An important
    feature of successful supervised machine learning applications is to be able to explain the predictions given by
    the regression or classification model 
    being used.
    However, most 
    state-of-the-art
    models that have good predictive power lead to predictions that are hard to interpret.
    Thus, several model-agnostic
    interpreters have been developed recently as a way of explaining black-box classifiers.
    In practice, using these methods
    is a slow process because a novel fitting is required for each new testing instance, and several non-trivial choices must be made.
    We develop NLS (neural local smoother), a method  that is complex enough to give good predictions, and yet gives solutions that are easy to be interpreted without the need of using a separate interpreter. The key idea is to use a neural network that imposes a local linear shape to the output layer. We show that NLS
    leads to predictive power that is comparable to state-of-the-art machine learning models, and yet is easier to interpret.
\end{abstract}
\end{@twocolumnfalse}
]

\clearpage

\section{Introduction}
\label{sec:intro}

Machine learning applications are often focused on maximizing prediction accuracy, leading practitioners to choose highly complex regression estimators \citep{Vach96neuralnetworks}. In this scenario, neural networks have recently gained much prominence in regression applications due to their high predictive accuracy and their scalability to large datasets \citep{lecun2015deep}. 

However, in many applications, accuracy is only one of the features that must be considered when choosing which prediction method to use. Another relevant aspect is the easiness in interpreting the outputs of the method at hand. The ability to explain predictions made by a method is important to give insights about the decisions being taken by the learned model, which can increase the trust practitioners have over the ML model \citep{doshi2017towards}.

In this work, we introduce the Neural Local Smoother (NLS), a one-step approach
to fit a neural network that yields predictions that are easy to be explained. The key idea of the method is to combine the architecture of the network with a local linear output  \citep{1906.09735}.
We show that while  NLS  keeps the high predictive accuracy of neural networks, it is highly interpretable.

\subsection{Related work}

Many approaches 
to interpreting complex machine learning algorithms have been proposed; see \citet{hechtlinger2016interpretation,koh2017understanding,lundberg2017unified,guidotti2019survey} and references therein for a review of some methods.
Typically, these proposals offer model agnostic interpreters to explain the outputs given by a learned model, that is, explanations to a prediction can be taken regardless of the model nature. An example of one solution is LIME \citep{ribeiro2016should}. LIME uses a kernel smoother to fit a local linear approximation to a (possibly complex) regression function around the instance to be explained. By looking at the coefficients of this approximation, it is then possible to explain why a particular prediction was made.

LIME and related methods work on demand. That is, every time a new instance, $\x^*$, needs to be explained, a local linear estimator is fit on a neighborhood around $\x^*$. This process can be too slow to be applied in practice. Moreover, several nontrivial choices on how to define the neighborhood around $\x^*$ and how to sample from it need to be made \citep{fong2017interpretable,botari2019}
Thus, this two-step approach of first learning the regression and then explaining it is not practical in many applications.
On the other hand, NLS learns a prediction model that is already locally linear and can, therefore, already be interpreted in the same way as an explanation given by LIME.


The remaining of the work is organized as follows: in Section \ref{sec:NLS}, we introduce NLS and show how it  outperforms other local linear approaches. In Section \ref{sec:app}, we apply the NLS to real data, comparing its performance to other state-of-the-art methods. Finally, Section \ref{sec:final} presents final remarks and possible future extensions.

\section{The Neuro Local Smoother - NLS}
\label{sec:NLS}

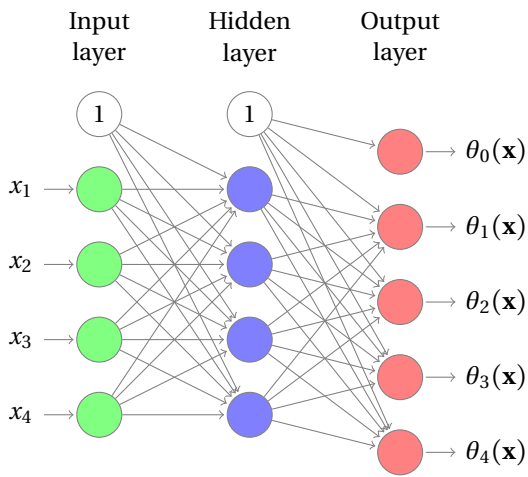
\begin{figure}[htbp]
\def\layersep{2cm}
\centering
\begin{tikzpicture}[shorten >=1pt,->,draw=black!50, node distance=\layersep]
    \tikzstyle{every pin edge}=[<-,shorten <=1pt];
    \tikzstyle{neuron}=[draw,circle,minimum size=17pt,inner sep=0pt];
    \tikzstyle{input neuron}=[neuron, fill=green!50];
    \tikzstyle{output neuron}=[neuron, fill=red!50];
    \tikzstyle{hidden neuron}=[neuron, fill=blue!50];
    \tikzstyle{annot} = [text width=4em, text centered];
   
    \node[neuron] (I-0) at (0,0) {1};
    \foreach \name / \y in {1,...,4}
        \node[input neuron, pin=left:$x_\name$] (I-\name) at (0,-\y) {};

    \node[neuron] (H-0) at (\layersep,0) {1};
    \foreach \name / \y in {1,...,4}
        \node[hidden neuron] (H-\name) at (\layersep,-\y) {};

    \node[output neuron,pin={[pin edge={->}]right:$\theta_{0}(\x)$}, right of=H-0] (O-0) at (\layersep,-0.5) {};
    \foreach \name / \y in {1,...,4}
        \path[yshift=-0.5cm]
            node[output neuron,pin={[pin edge={->}]right:$\theta_{\name}(\x)$}, right of=H-0] (O-\name) at (\layersep,-\y) {};

    \foreach \source in {0,...,4}
        \foreach \dest in {1,...,4}
            \path (I-\source) edge (H-\dest);

    \path (H-0) edge (O-0);
    \foreach \source in {0,...,4}
    \foreach \name in {1,...,4}
        \path (H-\source) edge (O-\name);

    \node[annot,above of=H-0, node distance=1cm] (hl) {Hidden layer};
    \node[annot,left of=hl] {Input layer};
    \node[annot,right of=hl] {Output layer};
\end{tikzpicture}
\caption{Example of a NLS neural network.}
\label{fig:exnet}
\end{figure}

Consider a set of data instances $(\X_1,Y_1), \ldots, (\X_n,Y_n)$, where $\X_i \in \mathbb{R}^d$ are features and $Y_i \in \mathbb{R}$ is the target to be predicted.
The Neural Local Smoother learns a neural network that ensures a local linear shape to the prediction function. 
In order to do so,
this neural network has input $\X$ and output $\Theta(\x) = (\theta_0, \theta_1(\x), \ldots, \theta_d(\x))$, where $d$ is the dimension of $\X$. 
An example of an NLS network containing 4 features and a single hidden layer with 4 neurons is shown in Figure \ref{fig:exnet}.
In order to obtain the predictions, these outputs are then combined according to
\begin{align}
\label{eq:nls}
    G_\Theta(\x) := \theta_0 +  \sum_{i = 1}^d \theta_i(\x) x_i.
\end{align} 
The prediction function of Equation \ref{eq:nls} is easy to interpret because it is locally linear.
Thus, given a new instance, $\x^*$, one can interpret the prediction
made to $\x^*$ by looking at the coefficients $\theta_i(\x^*)$, similar way as done in LIME.

Consider a fixed architecture of a NLS neural network that maps $\x \in \mathbf{R}^d$ into $\Theta(\x) \in \mathbf{R}^{d+1}$. Let $\Gamma$ be the set of all possible values for the parameters (weights) associated with that network. Each $\gamma \in \Gamma$ is then associated with a different choice of  $\Theta(\x)$. In order to learn the weights of the network, the NLS uses a squares loss function over a given training dataset $(\X_1, Y_1), \ldots, (\X_n, Y_n)$, that is,
\begin{align}
\gamma^* = \argmin_{\gamma \in \Gamma} \sum_{i=1}^n \left[(y_i - G_\Theta(\x_i))^2\right]
\label{eq:loss}
\end{align}
As long as the architecture
of the network is sufficiently complex, any regression function
can be represented by Equation \ref{eq:nls}. This property is confirmed by the following theorem: 
\begin{theorem}
\label{thm:NLS}
Let $r(\x):=\E[Y|\x]$ be the true regression function
and let $\epsilon>0$.
Assume that the domain of the feature space is $[0,1]^d$.
If $r(\x)$ is continuous, then there exists an architecture 
and weights for NLS such that
$|r(\x)-G_\Theta(\x)|<\epsilon$
for every $\x \in (0,1)^d$.
\end{theorem}

Theorem \ref{thm:NLS} implies that, for a complex enough architecture, an NLS can fully represent any neural network regression. Furthermore, for a fixed  index $i \in \{1, ...,d\}$, a NLS with $\theta_j(\x) \equiv 0 \, \forall j \neq i$ can still fully represent any neural network regression. Hence, there are infinite choices of $\Theta(\x)$, and thus infinite possible NLS fits that lead to the same predictions. In other words, the solution of Equation \ref{eq:loss} is not unique. There might be, therefore, a variety of $\gamma$ settings leading to similar predictive errors.

\subsection{Extending a local interpretation to its neighborhood}
A NLS is a local linear method, and thus $\theta_i(\x^*)$ can be locally interpreted as a linear coefficient. However, as \citet{ribeiro2018anchors} argue, practitioners tend to extend local interpretation to new samples, which can lead to poor inference if $\theta_i(\x)$ varies too much. 
Thus, to minimize this effect, $\theta_i(\x)$ should vary smoothly with $\x$.
Therefore, we define an alternative loss function that penalizes non smooth solutions through the cumulative squared derivative. We choose $\gamma$ as follows:
\begin{align}
\begin{aligned}
& \gamma^* = \argmin_{\gamma \in \Gamma} \sum_{i=1}^n \bigg[(y_i - G_\gamma(\x_i))^2
\\ & + \lambda \sum_{k,l\geq 0} \left( \frac{\partial \theta_k(\x)}{\partial x(l)}\Big|_{\x=\x_i} \right)^2 \bigg]
\end{aligned}
\label{eq:tradeoff}
\end{align}
where $\lambda$ is the penalization strength. This penalization  guarantees that the optimization algorithm pursues $\gamma$'s for which $\Theta$ is smooth, leading to more accurate inferences to new samples when interpretations are extended. 

Equation \ref{eq:tradeoff} establishes a global interpretability-accuracy trade-off. If $\lambda = 0$, $\Theta(\x)$ can vary freely, which typically leads to predictive models with better accuracy. As $\lambda \longrightarrow \infty$, we recover a standard least squares linear regression (i.e., constant $\theta_i$'s), which is highly interpretable, but has low predictive power in most cases. Notice, however, that high values of $\lambda$ encourage simpler NLS, which tends to increase model bias while decreasing its variance. Thus, accuracy may also increase with $\lambda$.
We discuss how $\lambda$ can be chosen in practice
in Section \ref{sec:penalization}. 

The approach of introducing a penalty that encourages explainability
in prediction methods has also been proposed by \citet{plumb2019regularizing} in a general framework. In our case, we choose a  regularizer that is particularly suitable to a neural network because it is easy to compute: the derivatives in Equation \ref{eq:tradeoff} come straight from the back-propagation algorithm on the network fit. 

Example \ref{ex:tradeoff} presents a toy experiment to show the interpretability-accuracy trade-off in practice.

\begin{example}
\label{ex:tradeoff}
In this example, we use a NLS to fit the function $y = \sin(x)$ in the interval $[0, 2\pi]$. For this, we sampled 2,000 points in this interval and fit the NLS for $\lambda$ varying in $[0, 1]$ using $80\%$ of the points (randomly selected). For the remaining $20\%$, we compute the MSE (Mean Squared Error) and the average squared gradient (in this case, as $x$ is uni-dimensional, this is the cumulative squared derivative) as a function of $\lambda$. Figure \ref{fig:tradeoff} illustrates the obtained results. For large $\lambda$, NLS leads to a simple linear regression with gradients close to zero; for small $\lambda$, the true regression function is better approximated.

\begin{figure}[htbp]
    \centering
    \includegraphics[width=\linewidth]{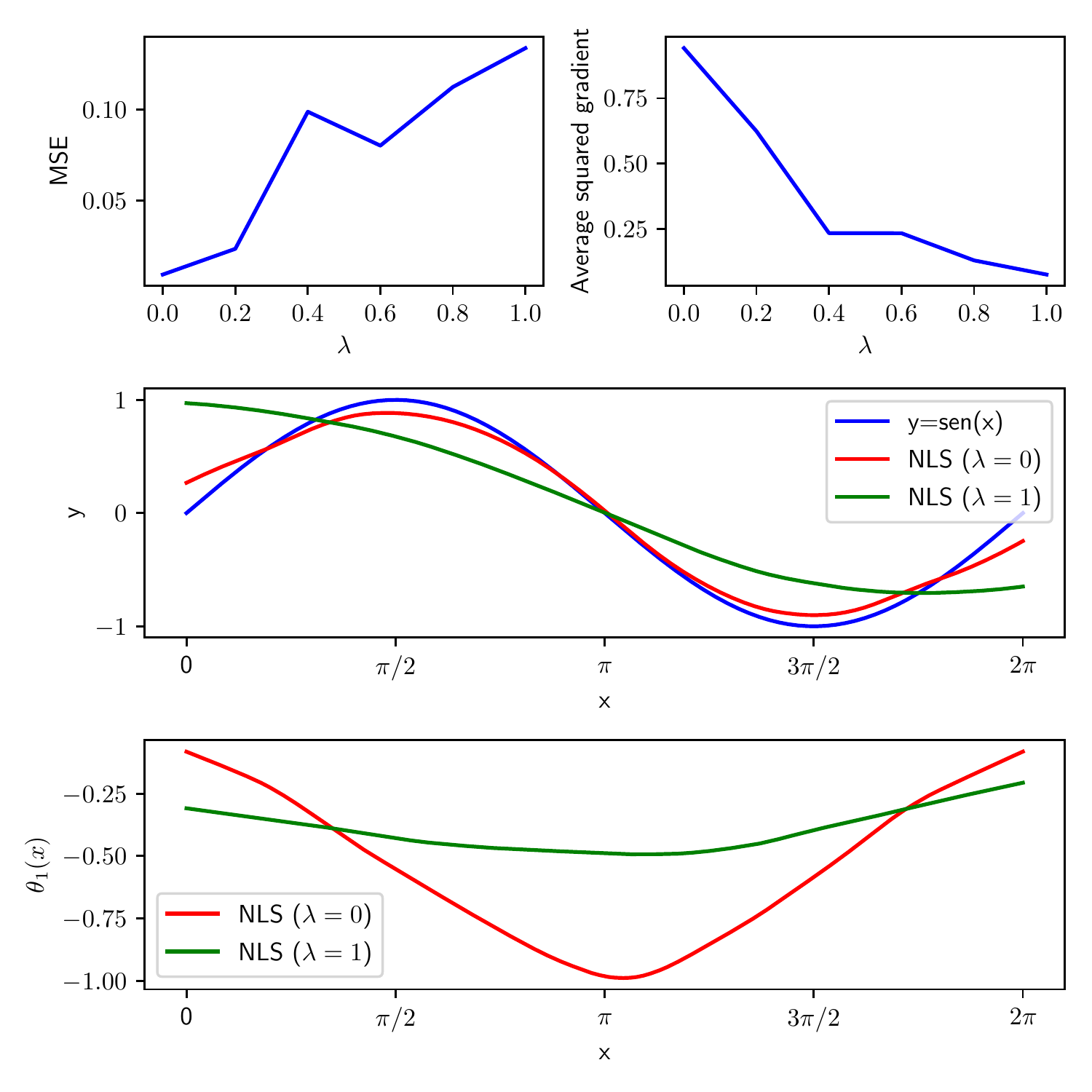}
    \caption{Top-left: MSE as a function of the penalization strength $\lambda$. Adding penalization decreases the NLS accuracy, illustrating the trade-off between accuracy and interpretability of the model. Top-right: Average squared gradient as a function of $\lambda$. High penalization values lead to smoother estimates for $\theta$. Middle and Bottom: True regression, NLS fits and fitted $\theta_1(x)$ for $\lambda = 0, 1$. While the non-penalized NLS yields a better fit, the penalized version provides a smoother estimate.}
    \label{fig:tradeoff}
\end{figure}
\end{example}


\subsection{Classification}

The NLS can also be generalized as a classifier. Assume that $Y$ assumes values in $\{0, 1, ..., L\}$.  Given an instance $\x$, the NLS estimates class probabilities with the shape 
\begin{align*}
    \P(Y = y | \x) \propto \exp\left(\theta_0^{(Y=y)} + \sum_{i=1}^d \theta^{(Y=y)}_i(\x) x_i\right).
\end{align*}
Hence, the classifier NLS neural network will estimate one vector $\Theta$ for each label $y$, that is, the network output has dimension $(k+1)(d+1)$. In order for $\P(Y = y | \x)$ to be well defined probabilities, we use the log-softmax function.
%
%
To optimize the network we use the cross entropy loss: we choose $\gamma$ via
\begin{align*}
    & \gamma^* = \argmin_{\gamma \in \Gamma} \sum_{i=1}^n \Big[ - \sum_{y=0}^L \mathbb{I}(Y_i = y)
     \log \P(Y_i = y_i | \x_i) \\ & + \lambda \sum_{k,l\geq 0} \left( \frac{\partial \theta_k(\x)}{\partial x(l)}\Big|_{\x=\x_i} \right)^2 \Big],
\end{align*}
where $\lambda$ controls the penalization strength as in the regression case. The logarithmic scale is used due to numeric optimization. The classifier NLS weights can be interpreted through the log-odds ratios as in standard logistic regression \citep{kutner2005applied}.

\subsection{Implementation details}
\label{sec:implementation}
The Python package that implements the methods proposed in this paper is available at
\url{github.com/randommm/nnlocallinear}.
We work with the following default specifications for the artificial neural networks:

\begin{itemize}
\item \textbf{Optimizer}: we choose to work with the Adam optimizer \citep{adam-optim} and decrease its learning rate once no improvement is achieved on the validation loss for a considerable number of epochs.

\item \textbf{Initialization}: to initialize the network weights, we used the method of initialization proposed by \citet{nn-initialization}.

\item \textbf{Layer activation and regularization}: we chose ELU \citep{elu} as the activation function and no regularization.

\item \textbf{Stop criterion}: a 90\%/10\% split early stopping for small datasets and a higher split factor for larger datasets (increasing the proportion of training instances) and a patience of 50 epochs without improvement on the validation set.

\item \textbf{Normalization}: batch normalization, as proposed by \citet{batch-normalization}, is used in this work in order to speed-up the training process.

\item \textbf{Dropout}: We also took advantage of dropout, which is a technique proposed by \citet{dropout}.

\item \textbf{Software}: we chose PyTorch as the deep learning framework.
\end{itemize}

\subsection{Connection to local linear estimators}

\label{sec:LLS}

 Local linear smoothing (LLS)
 methods \citep{fan1992variable,fan1992design,fan2018local} demonstrated to be a powerful tool for performing nonparametric regression in  several applications \citep{ruan2007real, mcmillen2004geographically}. Their prediction function has the shape
\begin{align}
\label{eq:local_linear}
    G_\Theta(\x) := \theta_0(\x) +  \sum_{i = 1}^d \theta_i(\x) x_i,
\end{align} 
that is, LLS also consists of a local linear expression for the regression function.
However, rather than estimating the parameters $\theta_i$ using neural networks,
 for each new instance $\x^*$,
 $\Theta(\x^*)=(\theta_0(\x^*),\ldots,\theta_d(\x^*))$ is estimated using  weighted least squares: 
\begin{align}
    \hat{\Theta}(\x) = \argmin_{\theta \in \mathbb{R}^{d+1}} \sum_{i=1}^n K(\x,\x_i)(Y_i-\theta_0 -  \sum_{i = 1}^d \theta_i x_i)^2,
\end{align}
\noindent
where $K$ is a smoothing kernel function. The solution to such a minimization problem is given by
\begin{align}
    \hat{\Theta}(\x^*) = (X^T W X)^{-1} X^T W y,
\end{align}
where $W = \mbox{diag}(K(\x^*, \x_1), K(\x^*, \x_2), ..., K(\x^*, \x_n))$.
Local linear smoothers yield good interpretability by default. One minor interpretability issue happens in LLS: On linear models, the parameter $\theta_0$ stands for $E[Y|x=0]$, but when $\theta_0(\x)$ is a function of $\x$, there is not a practical meaning for $\theta_0(\x)|\x \neq 0$. We fix this issue in NLS by fixing $\theta_0$.

The LLS calculates pairwise kernels for each new instance, leading to higher calculation effort and memory requirements as training data increases.
Also, a new least squares optimization is required
for each new instance. Therefore, local smoothers might be slow to generate predictions on high dimensional applications.
This is not the case for NLS: once the network is learned, evaluating the prediction on new instances only requires a single feed-forward run through the network.

In LLS, the kernel plays a key role: it controls the weight each training instance will receive, and thus  choosing a suitable kernel is important. In particular,
several algorithms to choose a suitable kernel have been developed
\citep{ali2006meta, khemchandani2009optimal, argyriou2006dc, hastie1993local}.
In practice, these methods require a family of kernel functions to be specified, such as a Gaussian kernel: \begin{align*}
    K(\x_i, \x_j) = \exp{\left\{-\frac{d^2(\x_i, \x_j)}{\sigma^2}\right\}}
\end{align*}
where $d(\cdot, \cdot)$ is the Euclidean distance and $\sigma^2$ a variance parameter that defines the size of the neighborhoods. $\sigma$ is usually chosen through cross-validation. Unfortunately, choosing a suitable kernel can be slow, especially because of the slow prediction time for LLS.

\begin{remark}
When using a Gaussian kernel on a local linear smoother, the $\sigma$ kernel hyper-parameter has a similar role as  $\lambda$ on NLS:  when $\sigma \longrightarrow \infty$, the weights are the same over the whole sample space, and hence the plain least squares linear regression is recovered. As $\sigma$ gets small, the local linear parameters can vary more.
\end{remark}

Notice that the Gaussian kernel (as well as the other most commonly used kernels) does not perform a weighting over the features, that is, every feature is equally relevant to the kernel value. In practice, however, features do not have the same predictive relevance: many of them can have no influence on $Y$. Therefore, an optimal weighting procedure should consider feature predictive relevance, which is not trivial. 
On the other hand,  NLS creates a local linear estimator that does not depend on a kernel. In particular, as the neural net has $\x$ as input, the network architecture automatically allows for feature selection. Example \ref{ex:simulation} shows a simulated example to illustrate this.
\vspace{2mm}
 
\begin{example}
\label{ex:simulation}
Consider the regression model $E(Y|x) = g(x) = x^2$. We sampled 2,000 instances with $x \sim U[-5, 5]$. We fitted a NLS (with
an architecture of 3 layers of size 500) and a LLS (with a Gaussian kernel, using cross-validation to choose $\sigma$). We also added irrelevant features (that is, features that are independent of the label) to the data and refitted the models. Figure \ref{fig:simulation} illustrates the features relationship with $Y$. Table \ref{tab:simulation} shows each model mean squared error on a $20\%$ holdout sample. 
The NLS is robust to these irrelevant features on the training data, while these can cause major damage to the accuracy of the LLS. 

\begin{figure}[htbp]
    \centering
    \includegraphics[width=\linewidth]{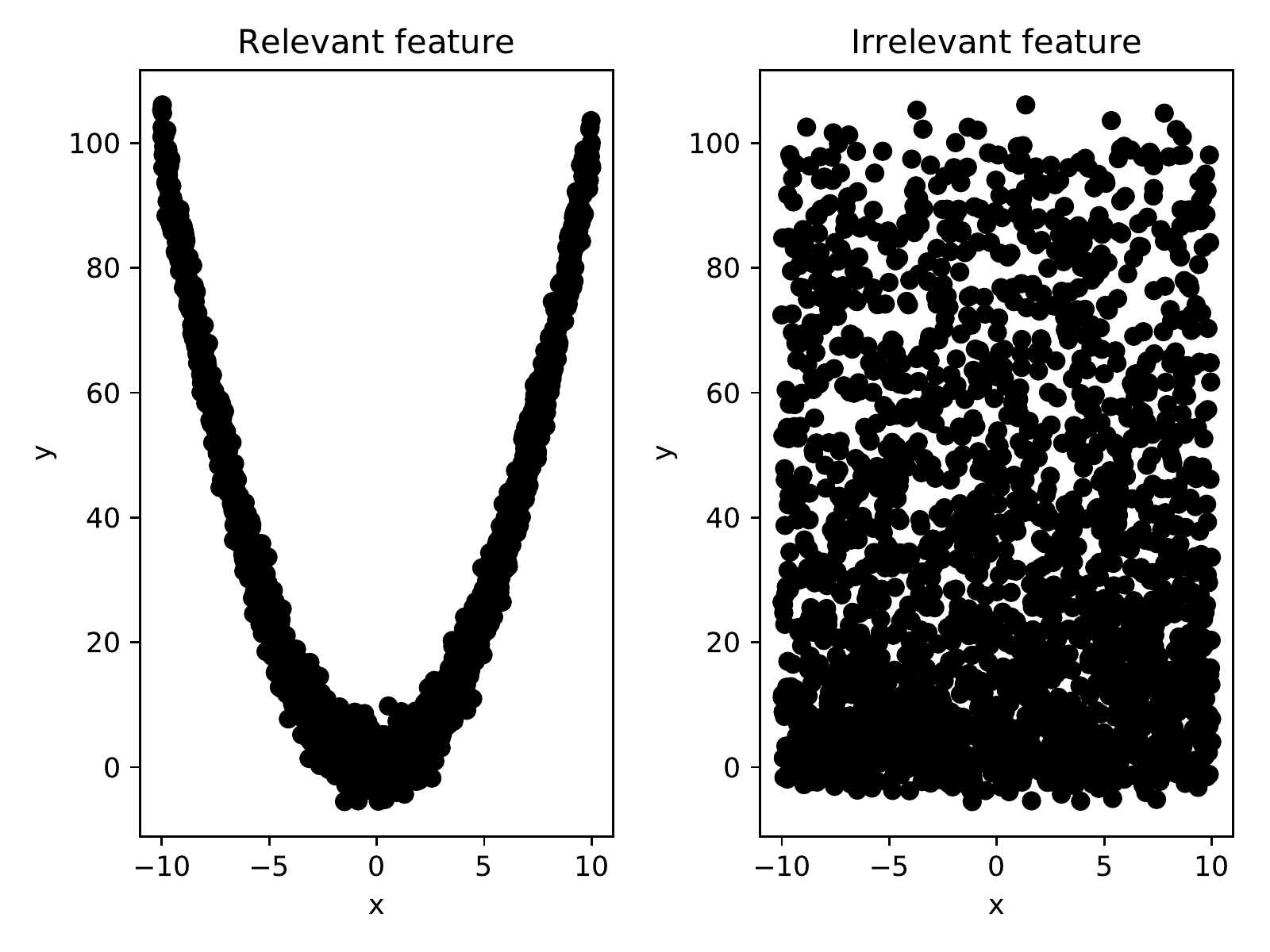}
    \caption{Relevant and irrelevant features relationship with $Y$. Although irrelevant features do not provide any information about $Y$,  kernels depend on all features, and therefore
    LLS does not perform well when they exist.}
    \label{fig:simulation}
\end{figure}

\begin{table*}[htbp]
    \centering
      \caption{Mean squared error of NLS and LLS  as a function of the number of irrelevant features. While LLS is heavily affected by irrelevant features due to the issues they cause on sample weighting, NLS
    does suffer as much.}
    \begin{tabular}{|c|c|c|c|} \hline
         \backslashbox{Model}{Irrelevant features} & 0 & 5 & 50 \\ \hline
         NLS & 8.64 & 8.83 & 13.89 \\ \hline
         LLS & 8.61 & 473.27 & 632.83 \\ \hline
    \end{tabular}
    \label{tab:simulation}
\end{table*}
\end{example}

\section{Experiments}
\label{sec:app}

The NLS is a prediction method that fits a local smoother through neural networks. Therefore, we want to ensure that it gives good predictive accuracy when compared to both standard neural network regression and LLS. In this section, we perform comparisons among these models, as well as  with random forests \citep{breiman2001random}. We used the following datasets:
\begin{itemize}
    \item The Boston housing dataset \citep{harrison1978hedonic} (506 instances, 13 features),
    \item The superconductivity dataset \citep{hamidieh2018data} (21.263 instances, 80 features),
    \item The blog feedback dataset \citep{buza2014feedback} (60.021 instances, 280 features),
    \item The Amazon fine foods dataset \citep{mcauley2013amateurs} (100.000 instances, textual data).
\end{itemize}

To evaluate models (on the test dataset), we used a split of $90\%$ for train-validation and $10\%$ test data for each dataset (except for the Boston dataset, for which we used 5-fold cross-validation procedure to split train-validation and test because of the small sample size). 
On each inner train-validation dataset, we used $90\%$ as train data and $10\%$ as validation data to perform a grid search optimized for the mean squared error (except for the Boston dataset, for which we used 2-fold cross-validation procedure to separate train and validation).
Moreover, for the neural network methods,  early stopping validation is performed on $10\%$  of the  training set. We describe below the grid search for each method:

\begin{itemize}
    \item For the NLS and the neural network regression (NN), we tested using 1, 3, and 5 layers, with sizes 100, 300 and 500 (9 combinations). We used no penalization for the NLS ($\lambda = 0$),
    \item For the LLS we used a Gaussian kernel and select the kernel variation parameter from $\{0.1, 1, 10, 100, 1000\}$,
    \item For the random forests (RF), we used the Scikit-learn \citep{scikit-learn} implementation and varied the number of trees in $\{10, 50, 100\}$.
\end{itemize}

For the final models obtained, we computed the MSE, the mean absolute error (MAE), and both metrics' standard deviations. Also, we evaluated the fitting time (in seconds) for every technique (including the cross-validation). These experiments were performed on an AMD Ryzen 7 1800X CPU running at 3.6Gz. Table \ref{tab:pp} shows the obtained results.

\begin{table*}[htbp]
\centering
\caption{MSE, MAE and their standard errors for the test set, and the fitting times.}
\begin{tabular}{|l|l|l|l|l|}
\hline
Data                               & Model & MSE                   & MAE                & Fitting time \\ \hline
\multirow{4}{*}{Boston housing}    & NLS   & 9.06   ($\pm$ 1.21)   & 2.11 ($\pm$ 0.10)  & 19500        \\ \cline{2-5} 
                                   & NN    & 12.31  ($\pm$ 1.95)   & 2.40 ($\pm$ 0.10)  & 17141        \\ \cline{2-5} 
                                   & LLS   & 11.82  ($\pm$ 1.86)   & 2.53 ($\pm$ 0.10)  & 2            \\ \cline{2-5} 
                                   & RF    & 8.90   ($\pm$ 1.40)   & 1.90 ($\pm$ 0.10)  & 1            \\ \hline
\multirow{4}{*}{Superconductivity} & NLS   & 105.50 ($\pm$ 8.13)   & 6.45 ($\pm$ 0.17)  & 4493         \\ \cline{2-5} 
                                   & NN    & 145.59 ($\pm$ 9.88)   & 7.81 ($\pm$ 0.20)  & 3610         \\ \cline{2-5} 
                                   & LLS   & 173.41 ($\pm$ 11.42)  & 8.29 ($\pm$ 0.22)  & 317          \\ \cline{2-5} 
                                   & RF    & 81.67  ($\pm$ 7.03)   & 5.04 ($\pm$ 0.16)  & 116          \\ \hline
\multirow{4}{*}{Blog feedback}     & NLS   & 271.23 ($\pm$ 40.72)  & 4.98 ($\pm$ 0.16)  & 14025        \\ \cline{2-5} 
                                   & NN    & 840.80 ($\pm$ 118.83) & 7.10 ($\pm$ 0.27)  & 38784        \\ \cline{2-5} 
                                   & LLS   & 273.81 ($\pm$ 47.67)  & 4.87 ($\pm$ 0.17)  & 3622         \\ \cline{2-5} 
                                   & RF    & 256.35 ($\pm$ 36.42)  & 3.34 ($\pm$ 0.15)  & 215          \\ \hline
\multirow{4}{*}{Amazon fine foods} & NLS   & 1.07 ($\pm$ 0.02)     & 0.69 ($\pm$ 0.01)  & 38754        \\ \cline{2-5} 
                                   & NN    & 1.14 ($\pm$ 0.02)     & 0.78 ($\pm$ 0.01)  & 121371       \\ \cline{2-5} 
                                   & LLS   & 1.06 ($\pm$ 0.01)     & 0.72 ($\pm$ 0.01)  & 6185         \\ \cline{2-5} 
                                   & RF    & 1.10 ($\pm$ 0.02)     & 0.70 ($\pm$ 0.01)  & 601          \\ \hline
\end{tabular}
\label{tab:pp}
\end{table*}

NLS either outperforms or has a similar performance against both LLS and NN in all of the datasets. When compared to random forests, NLS was the best in one out of four datasets.  As the NLS is estimated through neural networks, one could expect its performance to be poor on small training sets. However, the results for Boston housing data indicate that NLS can also lead to good predictions for small datasets. The training time of NLS is high (especially on high dimensional data), but, contrary to LLS, its predictions are fast to compute. 
There is also no need to fit an additional interpreter such as LIME to NLS on prediction time.

\subsection{Sample size effect and computational time}
\label{sec:amazon_sample_size}

\begin{figure}[htbp]
    \centering
    \includegraphics[width=\linewidth]{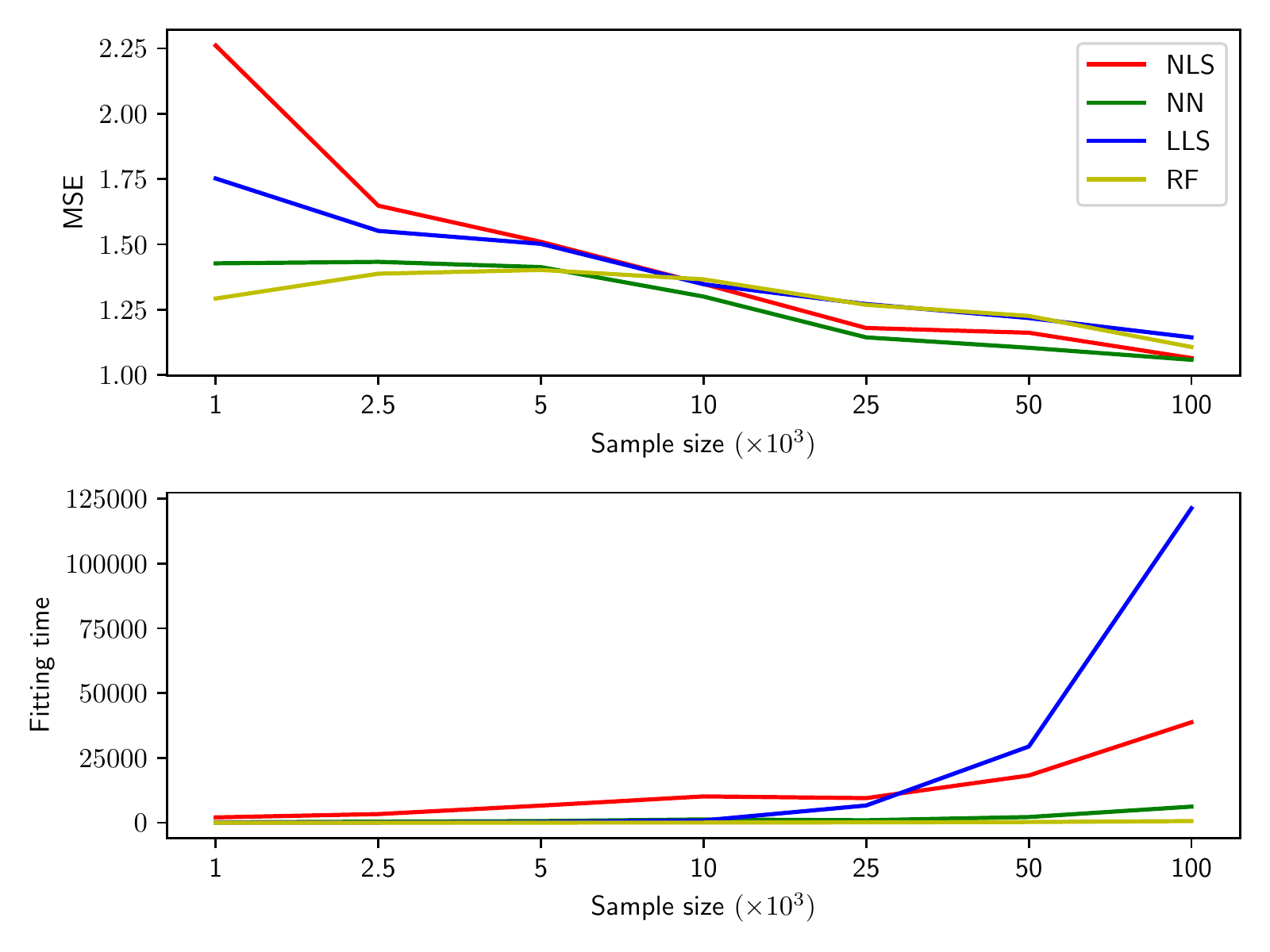}
    \caption{Top: Predictive accuracy over different sample sizes
    for the Amazon fine foods dataset. The NLS predictive performance increases in comparison to the others as the sample size increases. Bottom: Model fitting time for different sample sizes. While bigger samples massively increase the fitting time for the LLS, the NLS suffers a lower impact. Both methods are slow when compared to neural network regression and random forests. However, NLS is easy to interpret and is fast at prediction time.}
    \label{fig:amazon}
\end{figure}

Next, we use the Amazon fine foods dataset to check how the sample size affects both the quality and the fitting time of the models used in our previous experiments. 
In this experiment, we selected different sample sizes ranging from 1,000 to 100,000. For each sample size, we perform the same experiment described earlier in Section \ref{sec:app}. Figure \ref{fig:amazon} (top) shows how the test mean squared error of each method 
varies as a function of the sample size. While for smaller sample sizes, random forests give better predictions, NLS
becomes comparable to the other methods for $n>5000$.
Figure \ref{fig:amazon} (bottom) shows how the total fitting time (including cross-validation)  of each method varies as a function of the sample size. 
For small sample sizes, NLS is slow relative to the competing approaches, but for larger sample sizes, it becomes faster than the also easy-to-interpret LLS.

\subsection{The choice of \texorpdfstring{$\lambda$}{}}
\label{sec:penalization}

In this section we explore the role of the regularization parameter $\lambda$ from Equation \ref{eq:tradeoff}.

\begin{figure}[htbp]
    \centering
    \includegraphics[width=\linewidth]{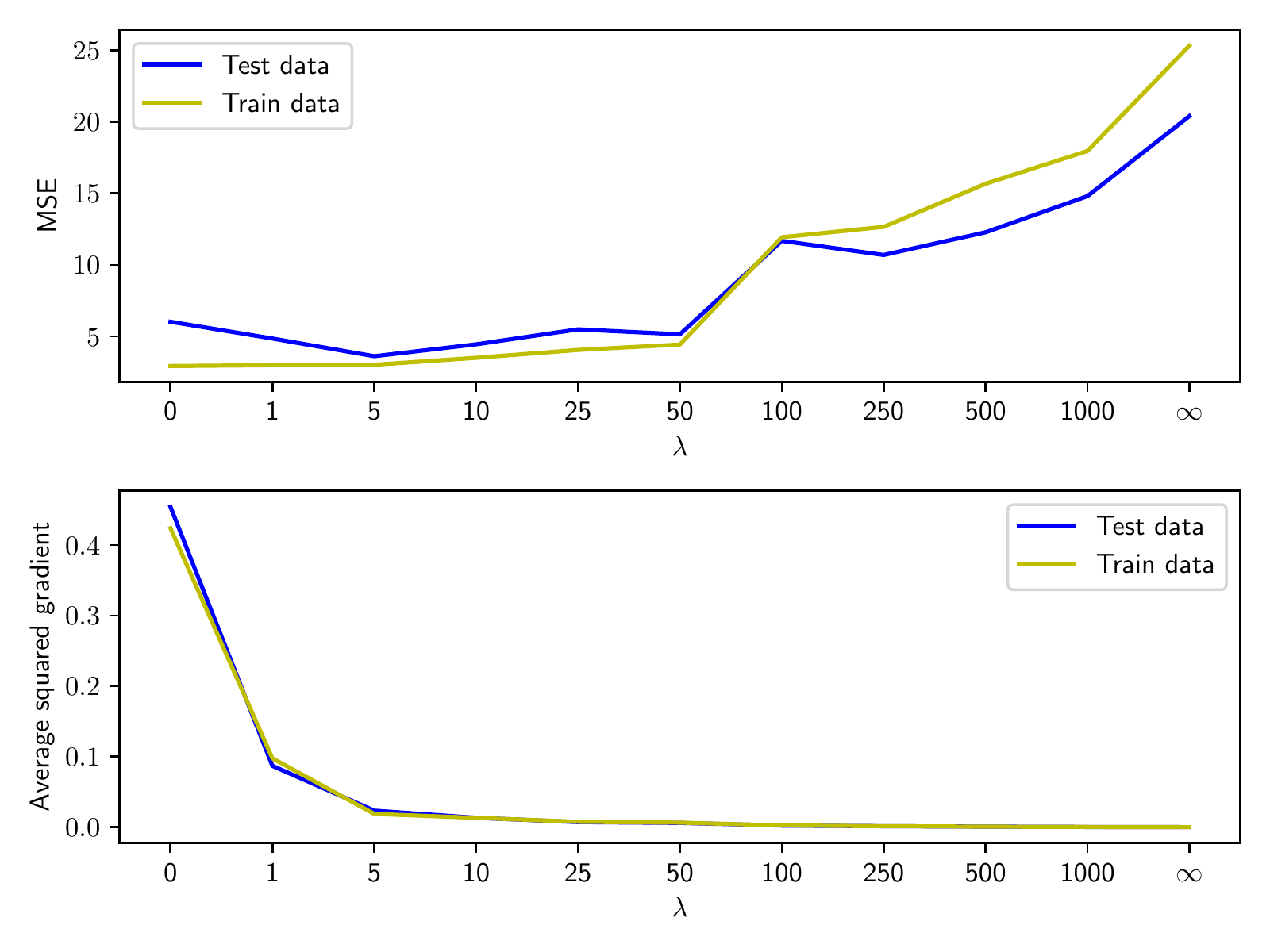}
    \caption{MSE and average squared gradients for both training and test datasets. Higher penalization values lead to smoother $\Theta(\x)$ estimates but higher train MSE.}
    \label{fig:lambda}
\end{figure}

In practice, to have an easy-to-interpret NLS that still yields good predictive performance, we successively increase $\lambda$ and check how the hold-out MSE varies. To reduce fitting time, for each $\lambda$ step, we initialize the network with the weights obtained in the last $\lambda$ that was used. 
We illustrate this procedure using the Boston housing dataset. We
start with the NLS fitted in Section \ref{sec:app} and refit it for $\lambda \in (0, \infty]$. Figure \ref{fig:lambda} illustrates how the  MSE and average squared gradient vary as a function of $\lambda$  both in the training and testing samples. The figure shows that an accuracy-interpretability trade-off occurs for the training data. On the other hand, in the test set, there is no big loss on increasing penalization factor $\lambda$ in the interval $[0, 50]$; in fact, increasing $\lambda$ can slightly decrease the MSE:
the fit with $\lambda = 5$ leads to the best test MSE value (3.61). An explanation for this fact is that the penalization  controls $\Theta(\x)$ variation, and thus controls over-fitting over the training data. Also notice that $\lambda = 5$ leads to substantially smoother solutions (bottom plot). Thus, in practice choosing $\lambda$ by data-splitting can give good prediction errors while maintaining an interpretable solution.

\subsection{NLS interpretation}

In machine learning, there is a debate about what  a good prediction explanation is \citep{doshi2017towards}. For example, \citet{ribeiro2018anchors} suggest that good explanations are the ones that allow human users to reproduce the regression function predictions for new samples with high accuracy after analyzing a set of given predictions and their explanations. In this section, we show how high penalization values allows users to reproduce NLS predictions. We use the Boston housing dataset as an example.

As $\lambda$ increases, we want to guarantee that NLS interpretations get more accurate in the sense proposed by \citet{ribeiro2018anchors}. That is, we want to guarantee that, given a set of predicted instances with their explanations, a naive algorithm can reproduce NLS predictions with high accuracy if $\lambda$ is large. 
 To test this statement, we use a set of given predictions and their explanations (i.e., the coefficients $\theta_i$ attached to them) to obtain predictions for an unseen instance $\x^e$ through a 1 nearest neighbor approach. We then compare such predictions to 
the true prediction for this instance,
$\mbox{true}\_{\mbox{pred}}(\x^e):=\hat{\theta}_0+\sum_{k=1}^d \hat{\theta}_k(\x^e)\x^k$.
Algorithm \ref{alg:lambda} describes the procedure for a fixed $\lambda$.

\begin{algorithm*}
  \caption{ \small Extending interpretations to replicate predictions}
  \label{alg:lambda}
  \algorithmicrequire \ {\small  
  A set of prediction instances $\{\x^p_1, ..., \x^p_n\}$, a set of extension instances $\{\x^e_1, ..., \x^e_n\}$.
  } \\
  \algorithmicensure \ {\small 
  Predictions obtained though interpretation extension for the extension instances.
  }

  \begin{algorithmic}[1]
  \For{$\x^e_i \in \{\x^e_1, ..., \x^e_n\}$}
    \State{Obtain $\x_{\mbox{neigh}} = \argmin_{\x^p_j \in \{\x^p_1, ..., \x^p_n)} d(\x^e_i, \x^p_j)$}
    \State{Obtain $\hat{\theta}_1(\x_{\mbox{neigh}}), ..., \hat{\theta}_d(\x_{\mbox{neigh}})$ through  NLS}
    \State{Evaluate $$\mbox{extended}\_{\mbox{pred}}(\x^e_i) := \hat{\theta}_0 + \sum_{k = 1}^d \hat{\theta}_k(\x_{\mbox{neigh}}) \x^e_{i, k}$$}
  \EndFor
  \end{algorithmic}
\end{algorithm*}

\begin{figure}[htbp]
    \centering
    \includegraphics[width=\linewidth]{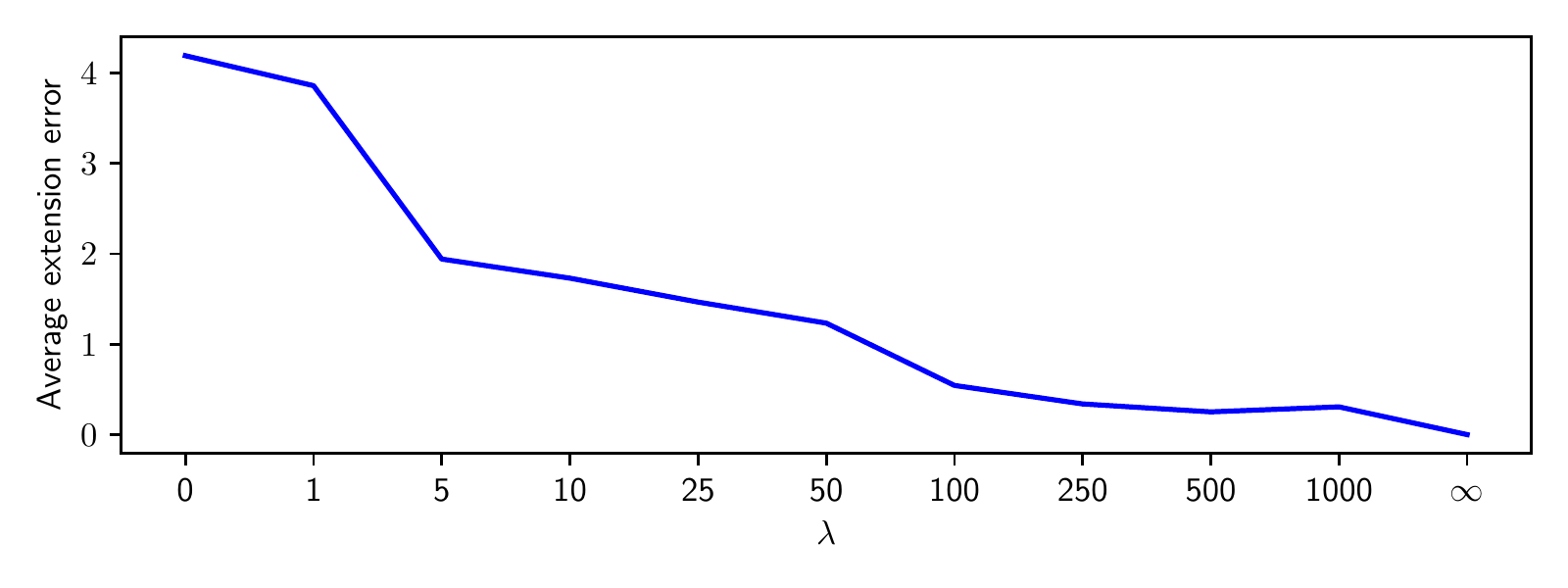}
    \caption{Average extension error as function of the penalization strength. Higher penalization values lead to more accurate prediction extensions.}
    \label{fig:extend}
\end{figure}

To ensure that extended predictions through interpretations are accurate, such predictions need to be compared to the ones given by the NLS. That is, we want to have low $|\mbox{extended}\_{\mbox{pred}}(\x^e_i) - \mbox{true}\_{\mbox{pred}}(\x^e_i)|$ on average. We split the test set ($3/4$ as prediction instances and the remaining as extension instances) and use Algorithm \ref{alg:lambda} for $\lambda$ values in $[2, \infty]$, and then compute such averages. Figure \ref{fig:extend} illustrates the results. The figure corroborates that the penalization strength $\lambda$ leads to an accuracy-interpretability trade-off
in the sense of
\citet{ribeiro2018anchors}.

\subsubsection{Comparison to LIME}
\begin{figure*}[htbp]
    \centering
    \includegraphics[width=\linewidth]{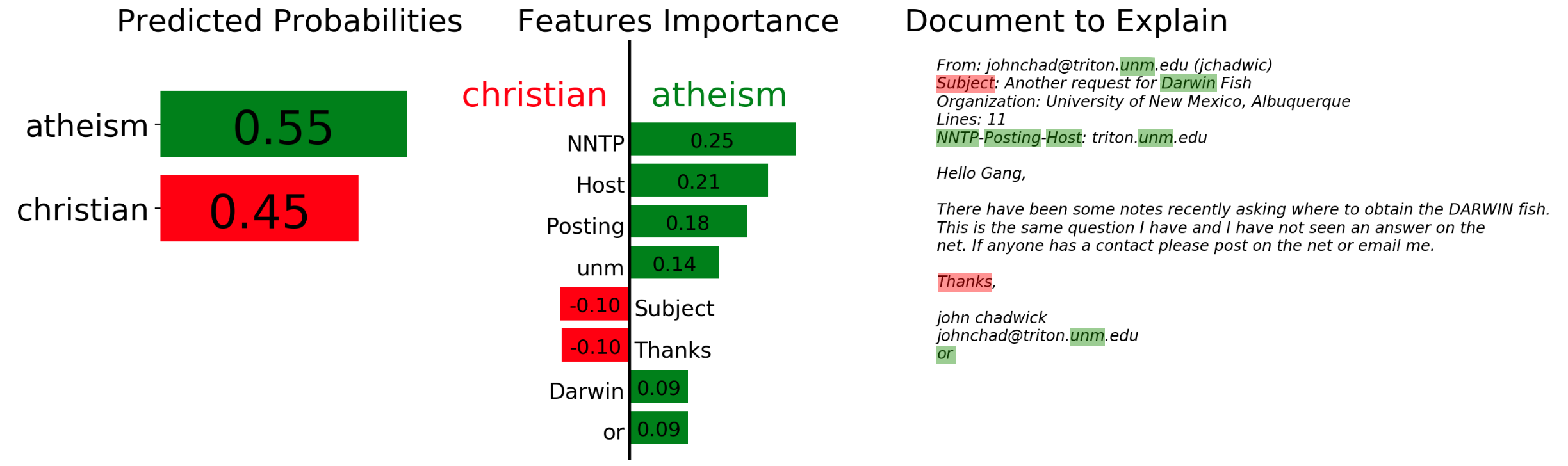}
    \caption{NLS predicted probabilities and explanations for the same instance investigated by \citet{ribeiro2016should}. NLS   has similar predictive power as SVM, and, at the same time, leads to interpretations that are similar to those obtained by LIME in this example.  }
    \label{fig:NLSexplain}
\end{figure*}

In this section, we apply  NLS to the same \emph{religion dataset} used to showcase LIME \citep{ribeiro2016should}. The goal is to verify if NLS leads to similar predictive power and captures similar explanations to the ones found in that paper. Figure \ref{fig:NLSexplain} shows the most relevant features that explain the predictions for this instance. These are words that appear in the mail header and not on its body. This finding is similar to the one obtained by applying LIME to a SVM model (see \citealt[Figure 2, right pannel]{ribeiro2016should}), and indicates that the dataset has issues. Furthermore, the accuracy obtained by NLS is 95,6\%, which is slightly larger than the one obtained by a SVM with RBF kernel (94\% according to \citealt{ribeiro2016should}). We conclude that for this example, the NLS (i) has slightly better predictive performance than SVM, and, at the same time, (ii) is able to give meaningful explanations.

\section{Conclusion and future work}
\label{sec:final}
 NLS is a prediction technique that enforces a local linear shape to a neural network. While NLS is able to represent the same functions as the usual predictive networks, it also allows users to make accurate interpretations directly using the estimated  coefficients, which are the output of the network. NLS presents some advantages when compared to local linear smoothers, as NLS is more robust to irrelevant features and generates predictions with almost no computation cost. We have also shown that NLS produces easy to interpret and accurate explanations for the given predictions in the sense proposed by 
\citep{ribeiro2018anchors}. Moreover, these explanations were comparable to those made by LIME in our experiments.

In future work, we wish to apply the NLS on classification datasets. 
Also, we will investigate  alternative loss functions for NLS that penalize non-sparse solutions. These alternative loss functions will ensure that the NLS is still highly interpretable on high dimensional applications.

\section*{Acknowledgments}
Victor Coscrato and Marco In\'{a}cio are grateful for the financial support of CAPES: this study was financed in part by the Coordena\c{c}\~{a}o de Aperfei\c{c}oamento de Pessoal de N\'{\i}vel Superior - Brasil (CAPES) - Finance Code 001. 
Rafael Izbicki is grateful for the financial support of FAPESP (grants 2017/03363-8 and 2019/11321-9) and CNPq (grant 306943/2017-4).
Tiago Botari acknowledges support by Grant 2017/06161-7, S\~ao Paulo Research Foundation (FAPESP).
This paper partially emanates from research supported by a grant from Science Foundation Ireland under Grant No. 18/CRT/6223 which is co-funded under the European Regional Development Fund.
The authors are also grateful for the suggestions given by Derek Bridge, Andr\'e C. P. L. F. Carvalho, Murilo C. Naldi, Marcos O. Prates, Diego F. Silva, and Rafael Stern.

\bibliography{main.bib}
\clearpage
\appendix

\section*{Appendix: proof}

\begin{proof}[Proof of Theorem \ref{thm:NLS}]

Because $h(\x):=\frac{r(\x)}{x_1}$
is also continuous,
it follows from the universal representation theorem \citep{cybenko1989approximations} that 
there exists a neural network with output $N(\x)$ such that 
$$| h(\x) -N(\x)|<\epsilon$$
for every $\x \in (0,1)^d$.

Now, let $\theta_1(\x)=N(\x)$,
$\theta_0=0$, and $\theta_i(\x)\equiv 0$ for every $i>1$, and thus
$G_\Theta(\x)=\theta_1(\x)x_1$.  Because $0 < x_1 < 1$, it follows that
$$|r(\x)-G_\Theta(\x)|=|r(\x)-N(\x)x_1| \leq$$ $$ | \frac{r(\x)}{x_1} -N(\x)|=|h(\x)-N(\x)|<\epsilon,$$ 
which concludes the proof.
\end{proof}
\end{document}